\documentclass[9pt]{article}

\usepackage[numbers]{natbib}
\usepackage{spconf}
\usepackage{amsmath,amssymb,amsfonts}
\usepackage{graphicx,color}
\usepackage{textcomp}
\usepackage{pifont}
\usepackage{xcolor}
\usepackage[hyphens]{url}
\usepackage{orcidlink}
\usepackage{hyperref}
\usepackage{bm}
\hypersetup{hidelinks}

\usepackage[numbers]{natbib}
\usepackage{algorithm,algorithmic}

\usepackage{enumitem}
\title{Advancing Newborn Care: Precise Birth Time Detection Using AI-Driven Thermal Imaging with Adaptive Normalization} 

\name{Jorge García-Torres$^{1}$ \qquad Øyvind Meinich-Bache$^{1,2}$ \qquad Anders Johannessen$^{2}$ \qquad Siren Rettedal$^{3,4}$ \qquad Vilde Kolstad$^{3}$ \qquad Kjersti Engan$^{1}$}

\address{$^{1}$ Dept. of Electrical Engineering and Computer Science, University of Stavanger, Norway \\
    $^{2}$ Laerdal Medical AS, Stavanger, Norway \\
    $^{3}$ Dept. for Simulation-based Learning, Stavanger University Hospital, Norway \\
    $^{4}$ Faculty of Health Sciences, University of Stavanger, Norway
    }

\begin{document}

\maketitle

\begin{abstract}
Around 5-10\% of newborns need assistance to start breathing. Currently, there is a lack of evidence-based research, objective data collection, and opportunities for learning from real newborn resuscitation emergency events. Generating and evaluating automated newborn resuscitation algorithm activity timelines relative to the Time of Birth (ToB) offers a promising opportunity to enhance newborn care practices. Given the importance of prompt resuscitation interventions within the "golden minute" after birth, having an accurate ToB with second precision is essential for effective subsequent analysis of newborn resuscitation episodes. Instead, ToB is generally registered manually, often with minute precision, making the process inefficient and susceptible to error and imprecision. In this work, we explore the fusion of Artificial Intelligence (AI) and thermal imaging to develop the first AI-driven ToB detector. The use of temperature information offers a promising alternative to detect the newborn while respecting the privacy of healthcare providers and mothers. However, the frequent inconsistencies in thermal measurements, especially in a multi-camera setup, make normalization strategies critical. Our methodology involves a three-step process: first, we propose an adaptive normalization method based on Gaussian mixture models (GMM) to mitigate issues related to temperature variations; second, we implement and deploy an AI model to detect the presence of the newborn within the thermal video frames; and third, we evaluate and post-process the model's predictions to estimate the ToB. A precision of 88.1\% and a recall of 89.3\% are reported in the detection of the newborn within thermal frames during performance evaluation. Our approach achieves an absolute median deviation of 2.7 seconds in estimating the ToB relative to the manual annotations.
\end{abstract}

% Keywords

% Main text
\section{Introduction}
\label{sec:intro}

\begin{figure*}[ht]
\centering
\begin{minipage}[b]{\linewidth}
  \centering
  \centerline{\includegraphics[width=\linewidth]{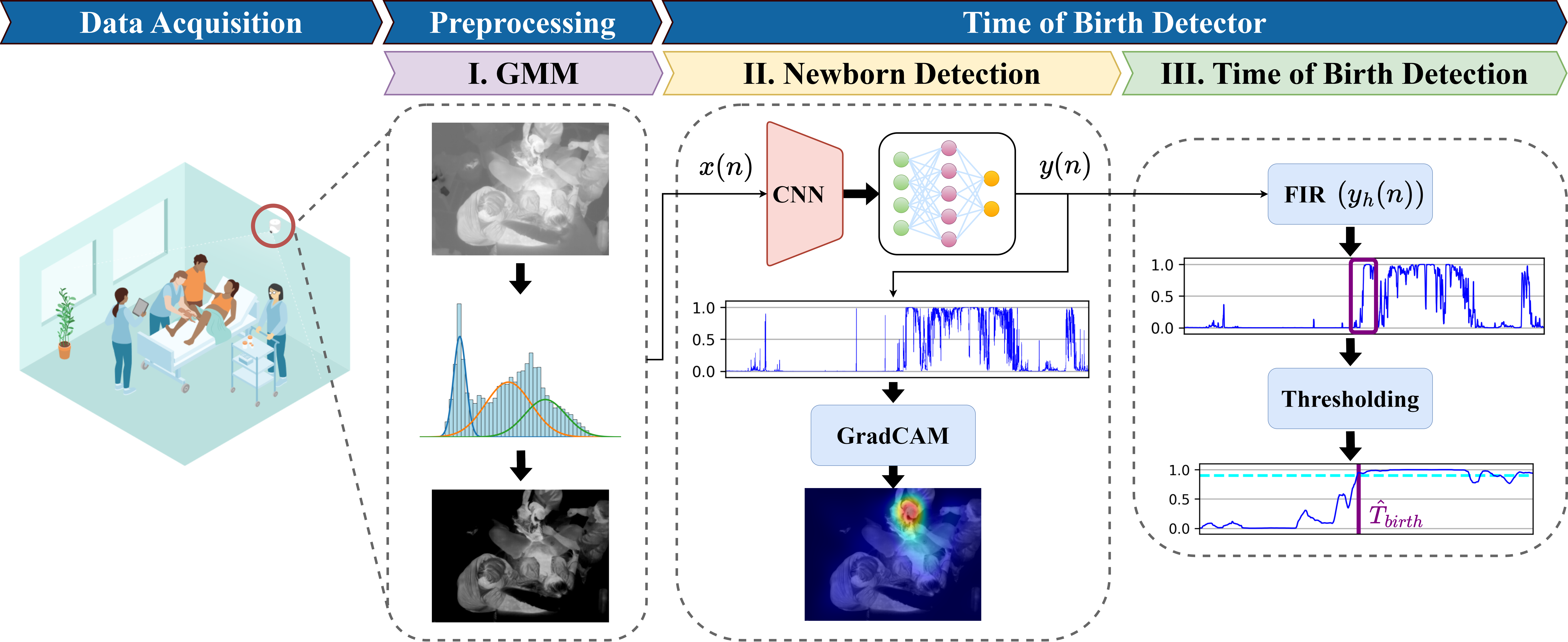}}
%  \vspace{1.5cm}
\end{minipage}
\caption{Overview of our proposed AI-driven Time of Birth detector. Birth episodes are captured using a thermal camera installed on the ceiling. Step I: our proposed Gaussian Mixture Model (GMM) normalization is employed in order to overcome the non-consistent distribution in the temperature values produced by other common normalization approaches. Step II: normalized frames are utilized as input $x(n)$ to our model to detect the presence of a newborn. We assess the model's ability to identify the newborn using GradCAM, providing insights into its decision-making process. Step III: the model's output scores $\hat{y}(n)$ are post-processed using a Finite Impulse Response (FIR) filter with a rectangular window. ToB is inferred based on the model's confidence in successfully detecting the newborn in individual video frames.}
\label{fig:overview}
\end{figure*}

Approximately 5-10\% of newborns experience insufficient breathing at birth and need some level of assistance to achieve cardiopulmonary stability \cite{ersdal2012early}. Birth asphyxia is one of the primary causes of neonatal mortality and morbidity, responsible for an estimated 900,000 deaths annually \cite{who-asphixia}. In accordance with neonatal resuscitation guidelines \cite{madar2021european, wyckoff20232022}, newborn resuscitation is time-critical, and immediately providing ventilation can significantly mitigate the risk of death and long-term damage related to birth asphyxia \cite{savechildren}. Effective resuscitative interventions should be initiated within the so-called ``golden minute''. This timeframe refers to the first minute after the Time of Birth (ToB), defined as the moment when the head, torso, and nates (gluteus) of the newborn become distinctly visible outside the mother’s perineum \cite{branche2020first}.

The analysis of newborn resuscitation videos has shown promising results for evaluation, debriefing, and training purposes \cite{skaare2018video, heydarzadeh2020impact}. Nevertheless, in order to accurately examine the impact of guidelines and determine the optimal timing and duration of the different Newborn Resuscitation Algorithm Activities (NRAA), a substantial number of documented NRAA episodes with detailed timeline information is essential. Manually annotating this data is time-consuming, inefficient, and raises privacy concerns. Besides, newborn resuscitation videos recorded at a resuscitation station lack pivotal information on the ToB and the duration of transferring newborns to the resuscitation station. To properly evaluate the treatment given, NRAA timelines must be relative to an accurate ToB, recorded with second precision. Currently, in clinical practice, ToB is recorded manually, often with minute precision, which is prone to error and imprecision. 

The potential of Artificial Intelligence (AI) has brought significant positive changes to the field of neonatology, improving neonatal applications such as imaging interpretation, prediction modeling using different health records, real-time monitoring integration, and documentation \cite{beam2023artificial, teji2022neoai, fischer2023end}. AI has also demonstrated the capability to automate NRAA timelines \cite{meinich2020activity, garcia-torres2023comparative}, showing promising outcomes for automated resuscitation video analysis. However, these timelines are dependent on manually annotated ToB. To the extent of our knowledge, no automated solution for detecting ToB exists. 

Thermal imaging enables the visualization of temperature variations by capturing the thermal radiation emitted by all the objects in a scene with a temperature above absolute zero (0 K, -273.15$^\circ$C). Albeit this technology has been applied to monitor human skin temperature for decades \cite{vardasca2019biomedical}, in the context of neonatal care, literature on thermal imaging remains scarce. Some studies have ventured into this area, and some scoping reviews can be found \cite{topalidou2019thermal}. Among others, it has been employed for assessing the health status of neonates \cite{ornek2019health}, identifying hypoxia immediately after birth \cite{urakova2017thermal}, measuring neonatal pulse rate with non-contact sensing \cite{paul2020non}, and monitoring respiratory rate \cite{maurya2023non}. However, employing thermal cameras presents challenges related to absolute temperature fluctuations due to several factors, including sensor noise, varying environmental conditions, and inconsistent calibration \cite{garcia2022towards}. This potential reliability issue is especially critical in applications involving multi-camera setups and where consistent temperature variations over time are crucial, requiring careful use of the absolute temperature values and highlighting the importance of finding an appropriate normalization process \cite{miethig2019convolutional}.

In this work, we aim to explore the combination of AI and thermal imaging in order to develop the first AI-driven ToB detector. Since the skin temperature of the newborn immediately after birth is higher than the skin temperature of other people in the room, thermal imaging can be used to detect the exact ToB while respecting the privacy of the healthcare personnel and the mothers. In this first automated ToB detection approach, we propose to utilize an image-based method, classifying frame by frame during inference to detect the presence of the newborn within the thermal frames. Prediction scores are then post-processed and used to estimate the ToB. Additionally, we introduce an adaptive normalization procedure based on Gaussian mixture models to address the challenges associated with utilizing thermal imaging. The block diagram of the proposed method is described in Figure~\ref{fig:overview}. This study is part of the NewbornTime project \cite{engan2023newborn}, an interdisciplinary collaboration that seeks to enhance newborn care by employing AI for activity and event recognition in videos during and after birth. The objective of this project is to develop NewbornTimeline, a system capable of automatically generating NRAA timelines, including the ToB and resuscitation activities. This tool is expected to facilitate the analysis of an extensive number of newborn resuscitation videos, contributing to a deeper understanding of optimal newborn resuscitation practices. 

Due to privacy concerns, ethical approvals, and data treatment policies, we are not allowed to share the sensitive data employed in this work. However, to compensate for this setback, we provide an implementation example with simulated data. The code is available here: \url{https://github.com/JTorres258/image-based-tob.git}.

\section{Data Material}
\label{sec:data}

\begin{figure}[t]
\centering
\begin{minipage}[b]{\linewidth}
  \centering
  \centerline{\includegraphics[width=\linewidth]{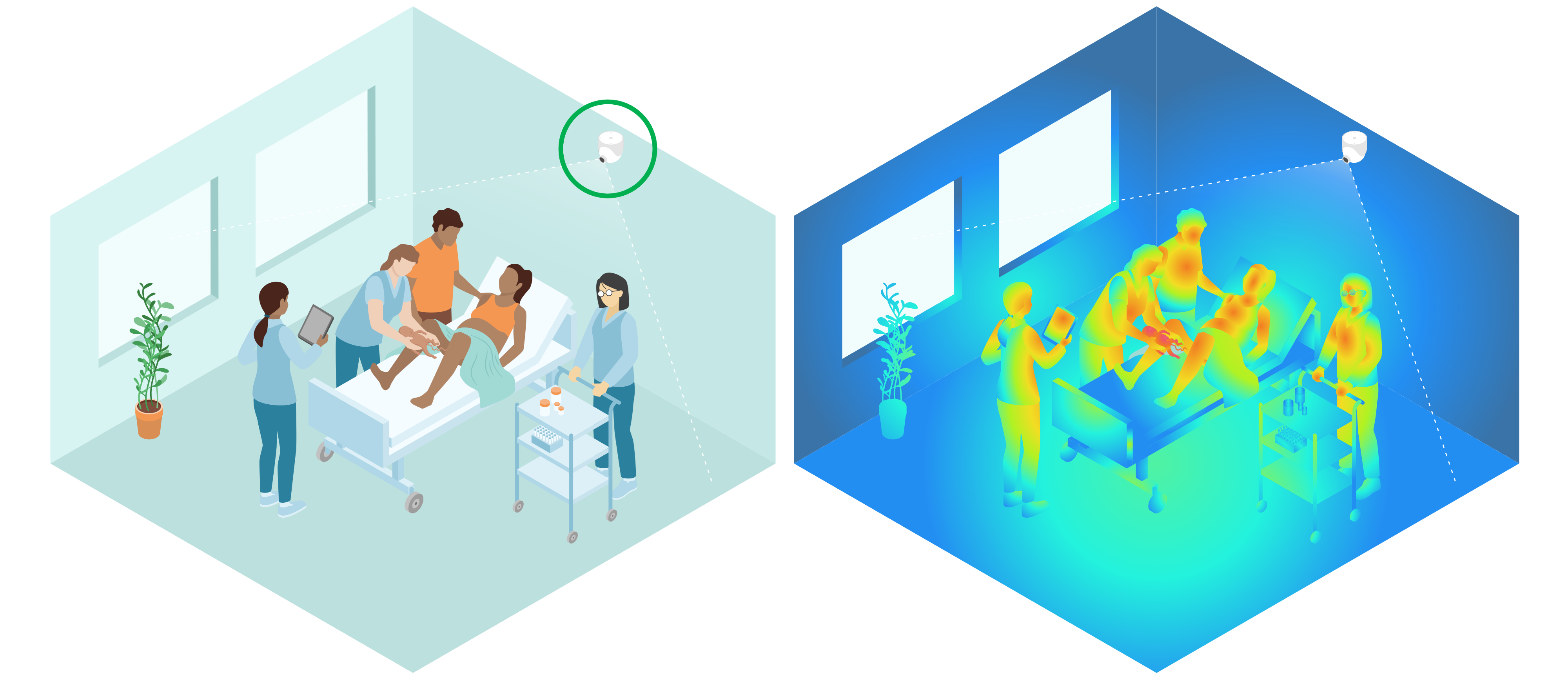}}
%  \vspace{1.5cm}
\end{minipage}
\caption{Illustration of the thermal module set-up in delivery rooms at SUS. On the left, the thermal sensor is mounted to the ceiling above the head of the mother, marked with a green circle. On the right, there is a representation of the thermal data from the birth scenario, with humans easily detectable due to their body temperature. The newborn, depicted in red, appears slightly warmer than the normal skin temperature at the time of birth. The illustration was created using Adobe Illustrator and published with permission from Laerdal Medical AS, attributed to Fredrik Kleppe.}
\label{fig:labour}
\end{figure}

Data were collected at Stavanger University Hospital (SUS), Norway, using a passive thermal module per room mounted in eight delivery rooms and one operation theatre. Each thermal module, provided by Mobotix \cite{mobotix:v1.02}, consisted of a thermal sensor Mx-O-SMA-TPR079 connected to an Mx-S16B camera. The sensor was mounted to the ceiling, centered above the head of the mother in the delivery rooms, as illustrated in Figure~\ref{fig:labour}. At the operating theatre, the thermal sensors are mounted to the ceiling minimizing blocking of the view from the operating light and other equipment. The dataset contains a total of 137 birth videos recorded with thermal cameras, with nearly equal distribution between delivery rooms and the operation theatre.

The data collection process was done semi-automatically. Thermal cameras initiated recording once a temperature value exceeded 30 degrees Celsius within a single pixel. This threshold was set at 30 degrees to streamline recording system operations, optimize data acquisition, and guarantee the detection of human presence. At delivery, ToB was manually registered by midwives or nurse assistants using a mobile application called Liveborn Observation App \cite{bucher2020digital} (Laerdal Medical AS, Stavanger, Norway). This application was specifically designed to document post-birth events in research projects and typically registers more accurate ToB than in standard clinical practice \cite{kolstad2024detection}. Following the birth registration in the Liveborn Observation App, a 30-minute thermal video was automatically produced by recording a period of $\pm$ 15 minutes centered around the manually registered ToB. Video frame resolution was 252$\times$336 with a frame rate of 8.33 frames per second. To retain thermal intensity information, video frames were saved in raw format, providing effectively 14 bits for data representation (values ranging from 0 to 16383). For a more intuitive representation, the intensity values can be converted into temperature values based on the specifications provided by Mobotix.

The annotation of birth videos was carried out by medical professionals using ELAN 5.8 (The Language Archive, Nijmegen, The Netherlands) \cite{wittenburg2006elan}. The dataset comprised various activity and event annotations across the thermal videos. For our study's specific objectives, our analysis centered on three labels: Visible Newborn (VNB), No Newborn (NNB), and ToB. VNB denotes the clear visibility and identification of the newborn within the video frame. NNB is deduced from the absence of VNB. ToB refers to the manually annotated Time of Birth relative to the thermal video (distinct from the manual registered ToB in LivebornApp during birth). In our dataset, 111 videos are fully annotated, whereas the remaining 26 videos only contain ToB information.

\section{Background \& Problem Formulation} % 
\label{sec:prep}

The skin temperature of a newborn immediately after birth is notably higher than the skin temperature of other people in the room. Effectively capturing and exploiting this information is crucial for the detection of ToB. In Kolstad et al. \cite{kolstad2024detection}, we demonstrated the capacity of thermal imaging to facilitate the recognition of ToB and cord clamping during manual visual inspections of birth videos captured by thermal cameras.

Despite these promising results, automating ToB detection is more challenging. In an ideal scenario where a single thermal camera is used, all thermal images would be captured under uniform conditions, with no variations in brightness or contrast. However, due to the inherent nature of thermal imaging and the fluctuating environmental conditions in a birth setting, maintaining consistency with the raw output of a thermal camera remains almost impossible. While most thermal imaging applications are either image-based, conducted in controlled environments with a single camera setup, or disregard the absolute temperature values, our study is different. We utilize multiple independent cameras without controlling the environment (to avoid interfering with medical professionals' work), but we focus on the sequential relationship between frames and the physical significance of relative temperature values.

To address the automated ToB detection, we unsuccessfully attempted to apply basic image processing techniques in a preliminary test using a few real births recorded using a thermal camera. Several environmental factors during birth had an adverse impact on the detection of ToB. The exposure of the newborn's skin could be brief, lasting only a few seconds before being given to the mother or covered with blankets. Healthcare providers usually practiced drying and stimulating the newborn during this time, potentially obstructing the thermal camera's view. Variations in the mother's birth position also added complexity to visualizing the newborn in the first seconds after the birth. Moreover, periods of time were observed where elements with similar temperatures or warmer than the newborn's skin temperature were present within the video frame, such as pots with hot water, wet towels, and blood. Additionally, we observed situations where the newborn was visible but was experiencing difficulty passing through the birth canal, not meeting the birth definition until fully delivered according to guidelines. All these factors are expected to occur regularly during birth episodes, increasing the level of difficulty in ToB detection.

\begin{figure}[t]
\centering
\begin{minipage}[b]{\linewidth}
  \centering
  \centerline{\includegraphics[width=\linewidth]{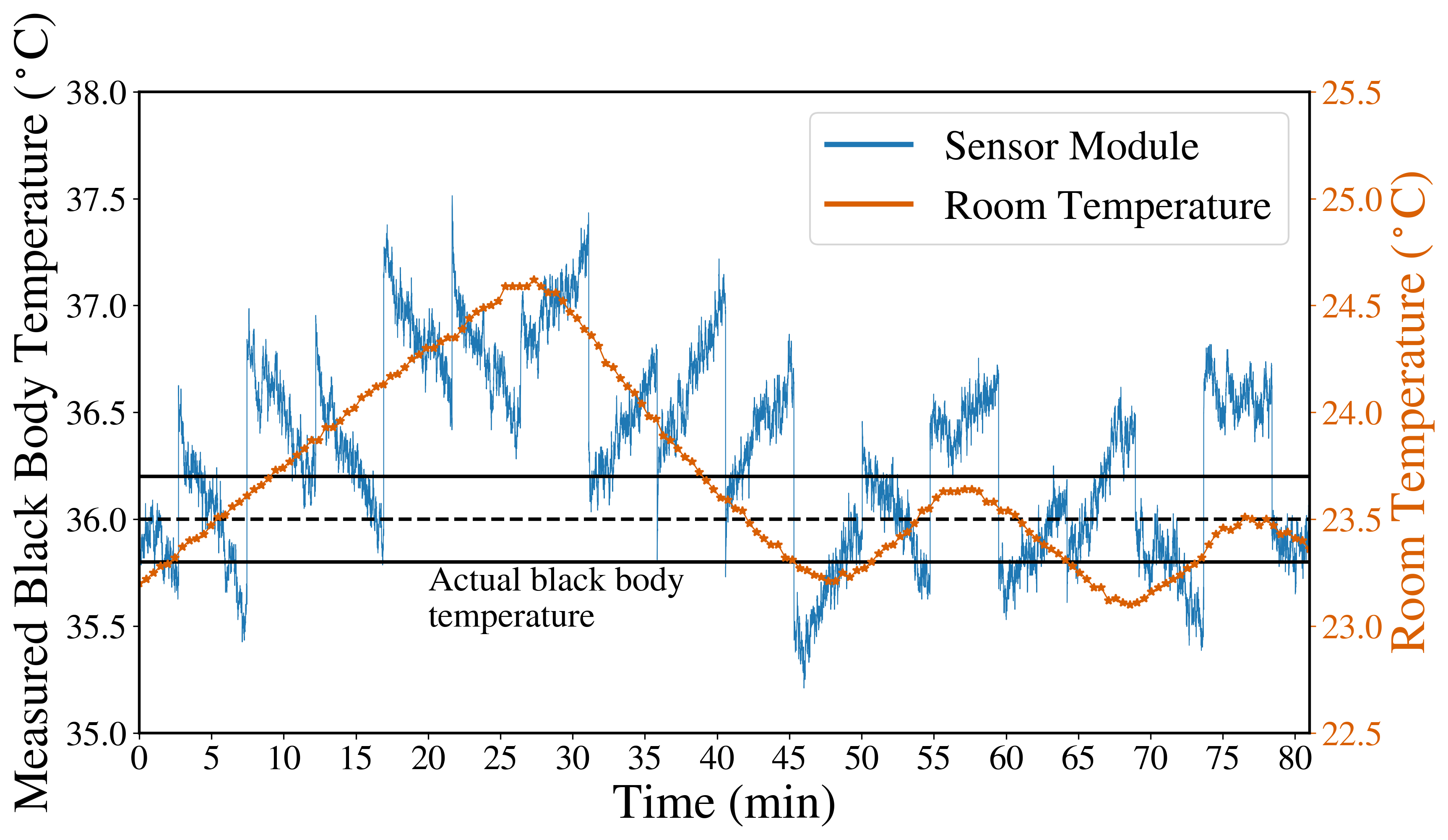}}
%  \vspace{1.5cm}
\end{minipage}
\caption{Temperature measurements on the surface of a black body device from the thermal sensor modules in a delivery room and the measured ambient room temperature along time. Variations in the measurements range between 35 and 38 degrees Celsius when the black body temperature is expected to be 36$\pm$0.2$^\circ$C. Illustration extracted from \cite{garcia2022towards} and modified.}
\label{fig:temp}
\end{figure}

\begin{figure}[t]
\centering
\begin{minipage}[b]{\linewidth}
  \centering
  \centerline{\includegraphics[width=\linewidth]{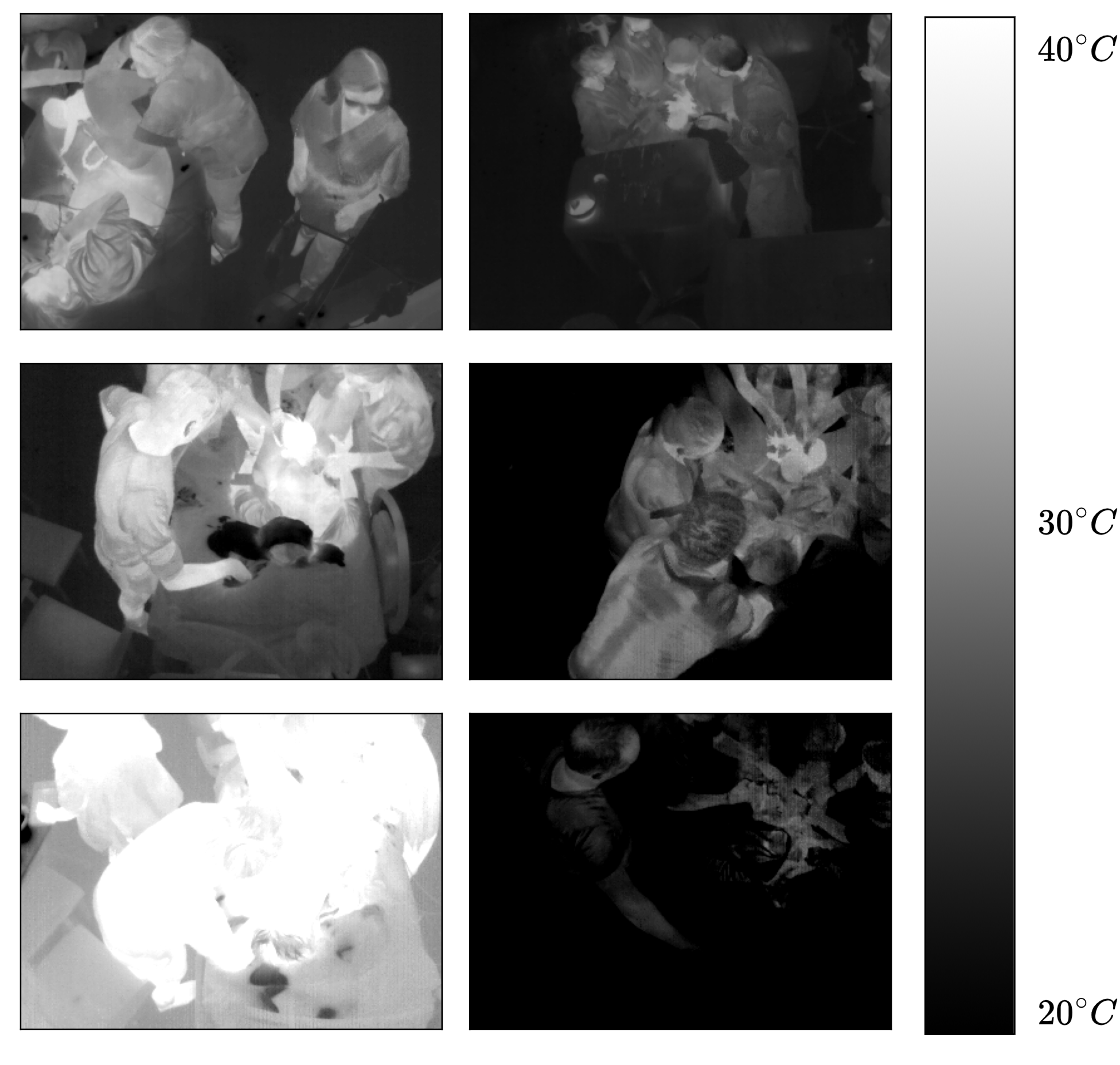}}
%  \vspace{1.5cm}
\end{minipage}
\caption{Illustration of raw thermal images. For visualization purposes, temperature values are clipped to a range of 20 to 40 degrees Celsius, representing the typical range of real-world temperatures for objects in the birth scenario. The first row shows the expected thermal visualizations, where human skin temperature is about 35$^\circ$ in delivery rooms and slightly lower in the operation theater. The second row highlights thermal images affected by temperature distortion factors, resulting in objects appearing either warmer (left, with newborn skin temperature exceeding 40$^\circ$) or cooler (right, with human skin temperature just above 30$^\circ$) than the expected temperature value. The third row demonstrates examples of thermal camera miscalibration, where temperature readings are significantly outside the expected range.}
\label{fig:raw_ex}
\end{figure}

Adversities were also identified in the utilization of thermal cameras, particularly in obtaining accurate temperature measurements for a specific object. In García-Torres et al. \cite{garcia2022towards}, we investigated the problems associated with using a thermal sensor setup for detecting ToB, where accurate temperature measurements would be desirable. Our investigation revealed that the temperature variations captured by thermal sensors included a combination of short-time random noise, drift, self-calibration mechanisms, and ambient room temperature, as shown in Figure~\ref{fig:temp}. Furthermore, thermal cameras can experience miscalibration over extended periods of time, requiring regular recalibration and maintenance. These distortions led to temperature disparities not only within a single thermal video but also across different videos, whether recorded with the same thermal camera or in different delivery rooms. Examples of raw thermal images from different videos are depicted in Figure~\ref{fig:raw_ex}. These findings highlight the limitations of relying on absolute temperature values for ToB detection, making thresholding or other simple image processing algorithms unfeasible for this task.

The necessity to rely on relative temperature values introduces a normalization challenge when deploying our AI system. Data normalization is a common technique to mitigate data skewness and enhance the learning curve of AI models. For grayscale or RGB images, where pixel values typically span from 0 to 255 in each channel, normalization involves adjusting intensity values to a smaller range, like 0 to 1 or -1 to 1. However, preserving the physical meaning of relative temperature values is crucial in thermal imaging applications like ours, where subtle temperature differences are significant. Applying a conventional Max-Min normalization approach to each thermal video frame based on their maximum and minimum values may cause the relative temperature difference compared to the other thermal frames to undergo a drastic alteration. Therefore, it is necessary to explore alternative methods that allow us to find an adaptive temperature range of interest within each thermal video. This targeted temperature range can then be employed to normalize across thermal frames, ensuring consistency.

\section{Methodology}
\label{sec:method}

In this paper, we present the first ToB detector based on AI and thermal imaging. As presented in Figure~\ref{fig:overview}, our methodology involves three steps. First, we propose an adaptive normalization method based on GMM to provide a more uniform representation of the data. Second, we formulate a binary image classification problem to detect the presence of the newborn within the thermal video frames using Convolutional Neural Networks (CNN). Third, we evaluate and post-process the model's predictions to estimate the ToB.

Going forward, let $\mathcal{V}\in\mathbb{R}^{N\times H\times W}$ denote a single-channel thermal video with $N$ thermal frames and a spatial resolution of $H\times W$. Let also $I(n)$ define the thermal frame at index $n=0,1,...,N-1$ so that:
\begin{equation}
    \mathcal{V} = \{I(0),I(1),...,I(N-1)\}
\end{equation}

\subsection{Step I. Gaussian Mixture Model Normalization}
\label{subsec:gmm}

The motivation behind the use of Gaussian Mixture Models (GMM) is to address the variability problem in temperature values. As outlined in Section~\ref{sec:prep}, some limitations make traditional normalization approaches unfeasible. Therefore, we aim to find a video-specific temperature range of interest that encompasses the human skin temperature. This adaptive approach enables us to normalize temperature values within the estimated range in each thermal video, focusing on the most relevant temperature variations for our study and generating more uniform data across videos.

\subsubsection{Gaussian Mixture Model}
\label{subsubsec:gmm_theory}

A Gaussian mixture model \cite{reynolds2009gaussian} is a parametric probabilistic density function represented as a weighted sum of \textit{M} Gaussian component densities. Let $\mathbf{z} \in R^{D}$ denote a continuous-valued data vector such as measurement or features. The Gaussian component densities, denoted as $g(\mathbf{z}|\bm{\mu}_i,\bm{\Sigma}_i),\,i=1,...,M$, collectively form the GMM probabilistic density function as follows:
\begin{equation}
    p(\mathbf{z}|\lambda)=\sum_{i=1}^M\alpha_ig(\mathbf{z}|\bm{\mu}_i,\bm{\Sigma}_i)
\end{equation}
where  $\alpha_i$ represents the mixture weights determining the contribution of each component, and $\lambda = \{\alpha_i, \bm{\mu}_i, \bm{\Sigma}_i\}$ is the set of parameters of the complete GMM. Each component density is a \textit{D}-variate Gaussian function given by:
\begin{equation}
    g(\mathbf{z}|\bm{\mu}_i,\bm{\Sigma}_i) = \\
    \frac{1}{(2\pi)^{D/2}|\bm{\Sigma}_i|^{1/2}}e^{\left(-\frac{1}{2}(\mathbf{z}-\bm{\mu}_i)'\bm{\Sigma}^{-1}_i(\mathbf{z}-\bm{\mu}_i)\right)}
\end{equation}
with mean vector $\bm{\mu}_i$ and covariance matrix $\bm{\Sigma}_i$. The mixture weights satisfy the constraint that $\sum_{i=1}^M\alpha_i=1$. For $\mathbf{z}$ denoting a 1D data vector, $\bm\mu_i$ is represented by a single value for each Gaussian component, $\mu_i$, and $\bm\Sigma_i$ is expressed as the variance $\sigma_i^2$.

The parameters $\lambda$ of the GMM are typically estimated using the Expectation-Maximization (EM) algorithm \cite{dempster1977maximum}. This iterative procedure comprises two main steps: the E-step (Expectation), involving the computation of the probability for each data point to belong to each cluster, and the M-step (Maximization), which updates the parameters based on these probabilities.

\subsubsection{Adaptive Normalization}
\label{subsubsec:gmm_norm}

\begin{figure}[t]
\centering
\begin{minipage}[b]{\linewidth}
  \centering
  \centerline{\includegraphics[width=\linewidth]{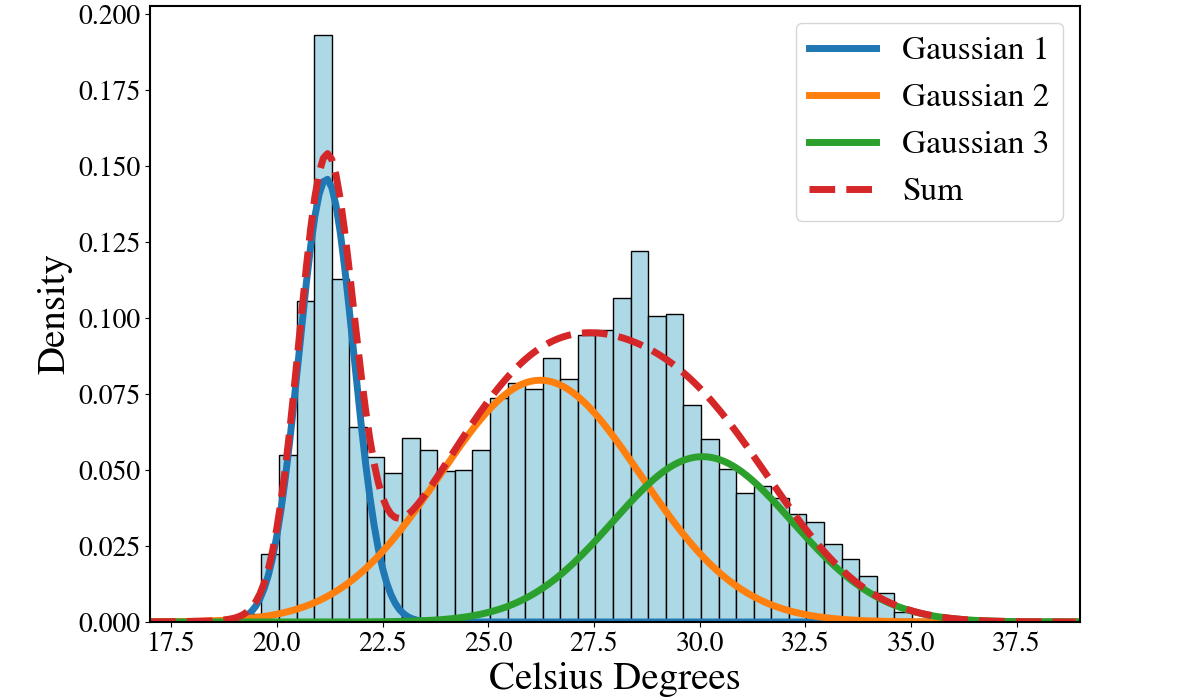}}
%  \vspace{1.5cm}
\end{minipage}
\caption{Representation of the complex data distribution of a thermal video and the density estimation provided by a GMM of three Gaussian components using temperature values from a thermal video. In blue, orange, and green, the three GMM components extracted from the data distribution. In red, the cumulative sum of all the GMM components.}
\label{fig:gmm}
\end{figure}

\begin{figure*}[t]
\centering
\begin{minipage}[b]{\linewidth}
  \centering
  \centerline{\includegraphics[width=\linewidth]{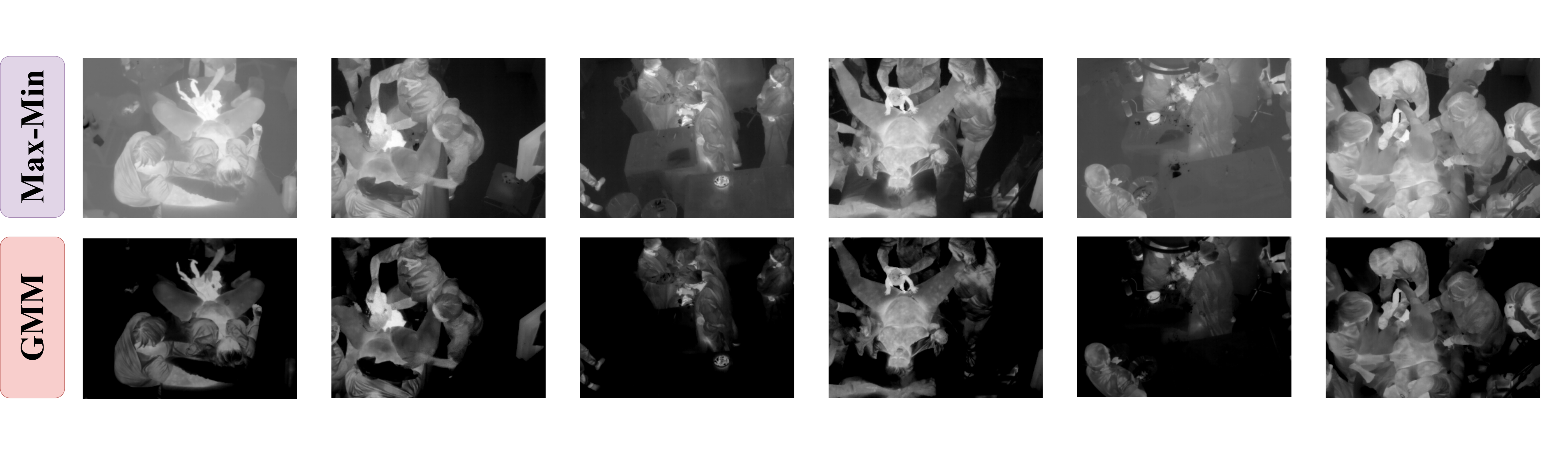}}
%  \vspace{1.5cm}
\end{minipage}
\caption{Comparison of the visual outcomes of Max-Min normalization and our proposed GMM normalization. The visual representation demonstrates the distinctive effects of each normalization method on temperature values, highlighting the advantages of GMM normalization in achieving a more non-saturated and consistent distribution across the thermal images within the same video and across multiple thermal videos.}
\label{fig:gmm_ex}
\end{figure*}

We applied GMM under the assumption that thermal data can be effectively represented within three primary regions: low temperatures (background), mid temperatures (clothes, hair, bed sheet, etc), and high temperatures (human skin and very warm elements). Therefore, $M=3$ Gaussian components are used to model the density distribution of the thermal data. Figure~\ref{fig:gmm} provides an example of generating a GMM using a thermal video from our dataset. For each thermal video $\mathcal{V}$, intensity values are extracted from frames spaced at 30-second intervals to capture a comprehensive representation. These values are converted into temperature values and organized into a one-dimensional temperature value vector $\mathbf{v}$, which is then used to fit the GMM $p(\mathbf{v}|\lambda)$. Among these components, the one with the highest mean value, denoted as $\hat{\mu_\mathbf{v}}=max(\mu_i)$, is selected. To prevent the selection of Gaussian components with a very wide distribution or an insignificant weight, the following empirical constraint is imposed based on the variance $\sigma_\mathbf{v}^2$ of the temperature values:
\begin{equation}
    \sigma^2_i \leq 2\sigma_\mathbf{v}^2 \textbf{  and  } \alpha_i \geq 0.15
\end{equation}
\begin{figure}[t]
\centering
\begin{minipage}[b]{\linewidth}
  \centering
  \centerline{\includegraphics[width=\linewidth]{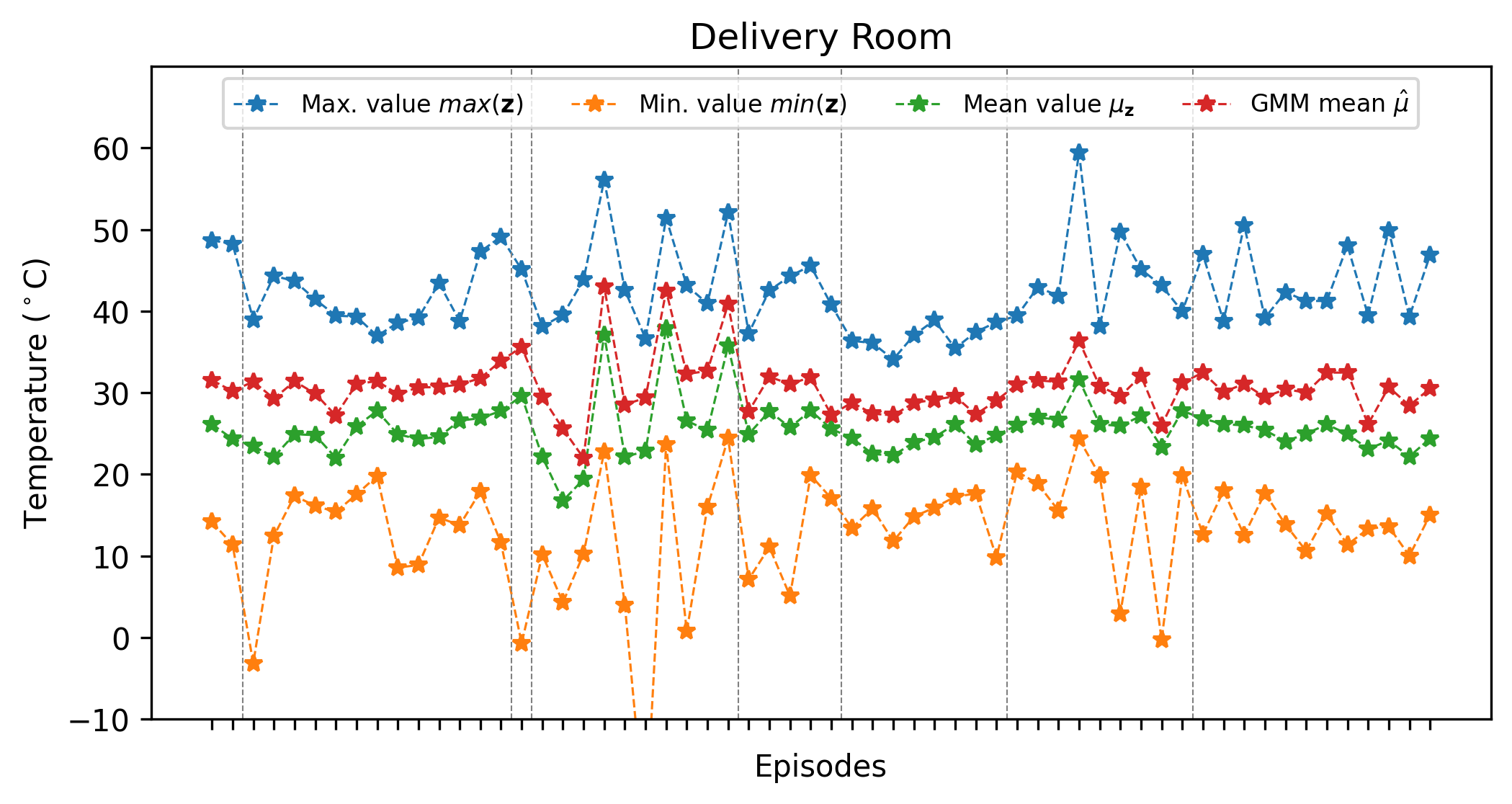}}
%  \vspace{1.5cm}
\end{minipage}
\begin{minipage}[b]{\linewidth}
  \centering
  \centerline{\includegraphics[width=\linewidth]{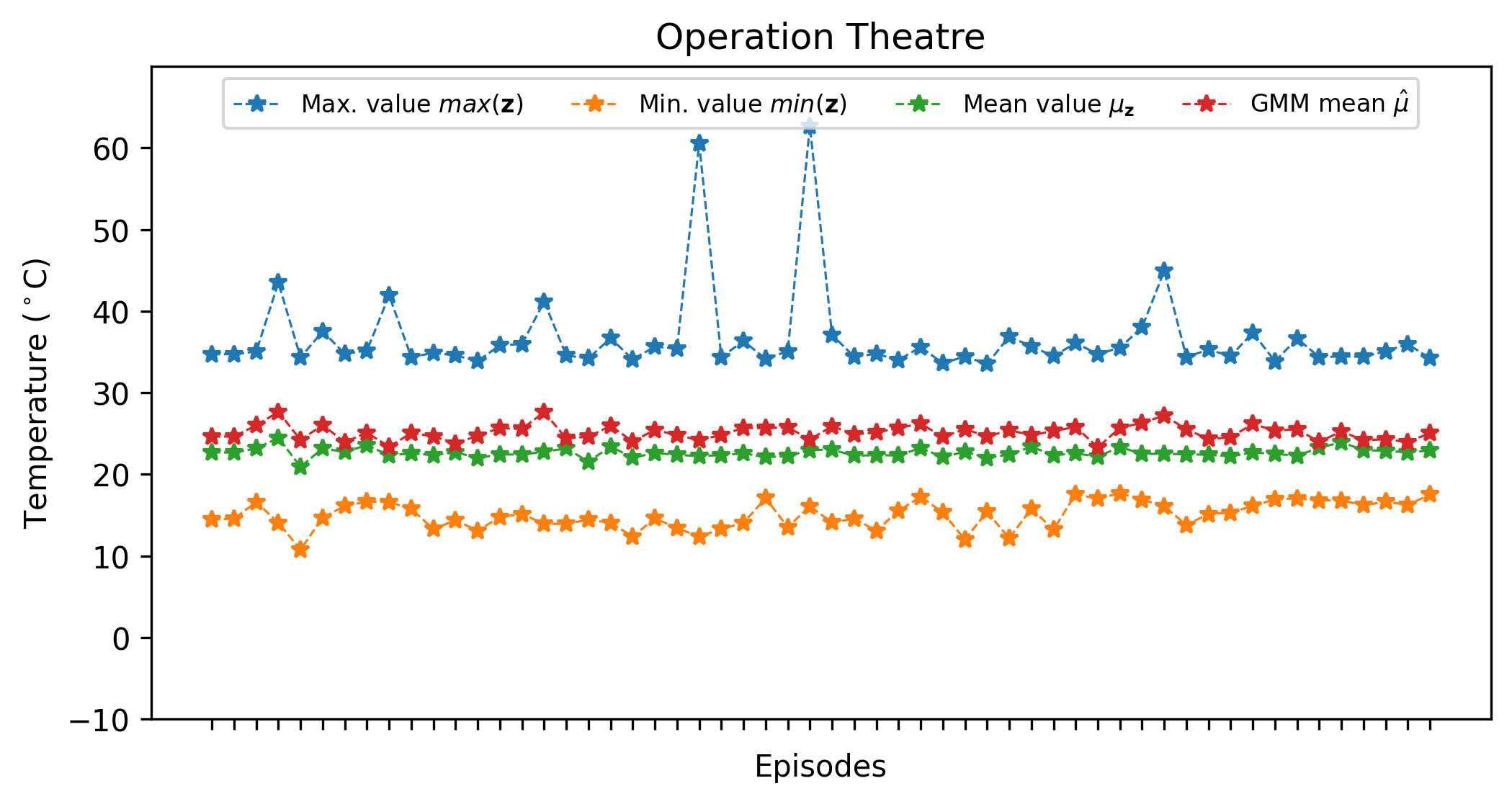}}
%  \vspace{1.5cm}
\end{minipage}
\caption{Evaluation of the temperature values used to fit the GMM for individual thermal videos. The maximum (blue), the minimum (orange), and the mean (green) temperature values are represented. In red, the highest mean temperature value selected from the GMM. Gray vertical lines separate episodes within the same delivery room, indicating temperature fluctuations even within the same room. In the operation theatre, temperature values show fewer variations across videos compared to the delivery rooms, likely attributed to more consistent ambient temperature conditions. No gray lines are represented as there is only one operation room for deliveries.}
\label{fig:gmm_eval}
\end{figure}
Once $\hat{\mu_\mathbf{v}}$ is determined, we utilize this value to define our temperature range of interest. Since variations in the temperature values are expected to happen, as depicted in Figure~\ref{fig:temp}, it is necessary to establish a sufficiently wide range to ensure that the skin temperature of the persons in the room and the newborn is always captured. From Figure~\ref{fig:gmm_eval}, we can observe that $\hat{\mu_\mathbf{v}}$ falls between the mean value $\mu_\mathbf{v}$ and $max(\mathbf{v})$ across all thermal videos. We use this information to set the lower bound of our range of interest by assessing the temperature difference between $\mu_\mathbf{v}$ and $\hat{\mu_\mathbf{v}}$ for each video and then computing the median. This results in a rounded lower value of -2.5$^\circ$C in the operation theatre and -5$^\circ$C in the delivery rooms. The upper bound is determined similarly but calculating the temperature difference between $max(\mathbf{v})$ and $\hat{\mu_\mathbf{v}}$ for each video, giving a rounded median value of 10$^\circ$C in both cases. However, to encompass the same span of temperature values in both scenarios, we extend the upper bound in the operation theatre to 12.5$^\circ$C. Therefore, the range of interest is defined as $[\hat{\mu_\mathbf{v}}-5,\hat{\mu_\mathbf{v}}+10]$$^\circ$C in the delivery rooms and $[\hat{\mu_\mathbf{v}}-2.5, \hat{\mu_\mathbf{v}}+12.5]$$^\circ$C in the operation theatre. Finally, temperature values are clipped within the aforementioned ranges and then rescaled to fall between 0 and 1, resulting in a normalized thermal video $\Tilde{\mathcal{V}} = \phi_{\hat{\mu}_\mathbf{v}}(\mathcal{V})$ where $\phi_{\hat{\mu}_\mathbf{v}}$ represents the GMM normalization function. Figure~\ref{fig:gmm_ex} shows examples of GMM and Max-Min normalization from six different thermal videos, highlighting the more non-saturated and consistent values provided by GMM normalization.

\subsection{Step II. Newborn Detection}
\label{sec:newborn_detection}

\subsubsection{CNN Backbone}
\label{subsubsec:arch}

Parameters optimization, computational efficiency, inference speed, and accuracy are specific requirements when designing CNN architectures for lightweight model deployments, particularly on resource-constrained devices. Considering the future implementation of our ToB detector in a real-time system in settings with limited computational resources, we prioritize the selection of CNN backbones for feature extraction that achieves a careful balance between accuracy and efficiency. Utilizing existing architectures also allows us to use transfer learning from pre-trained weights. In particular, we use MobileNetV3 \cite{howard2019searching} and EfficientNetV2 \cite{tan2021efficientnetv2}, the latest iterations of MobileNet and EfficientNet. Since multiple variations of these models exist, we focus on the smallest and largest versions of MobileNetV3 (Small and Large) and the base versions of EfficientNetV2 (B0, B1, B2, and B3) for comparison. 

For each backbone, we adjust its input size to match our thermal video frame resolution. We also replace the original architecture's classifier layer with one classifier containing two fully connected layers with 1024 and 2 units (binary classification), respectively, with randomized weights and a dropout layer of 0.5 in between during training. When pre-trained weights are used, the single-channel input data is adapted to the same number of channels as RGB data. This is achieved by expanding the intensity (grayscale) values to the three channels. Otherwise, backbone layers are adapted to match the single-channel input shape. In both cases, as implemented in MobileNetV3 and EfficientNetV2, input values are scaled to fall between -1 and 1 prior to being fed into the network.

\subsubsection{Implementation Details}
\label{subsubsec:setup}

Binary cross-entropy \cite{bishop2006pattern} is employed as the loss function. Since it is a binary classification problem, we define the positive prediction score for newborn detection as $\hat{y}$ for simplicity. To mitigate the impact of class imbalance during the model training, we estimate the inverted class weight $w_c$ for each class $c\in\{0 \text{ (NB)},1 \text{ (ToB)}\}$. Let $S$ denote the total number of samples in the training set, $C$ the number of classes, and $s_c$ the number of instances for a particular class $c$. The inverted class weight is defined as follows:
\begin{equation}
    w_c = \frac{S}{Cs_c}
\end{equation}
With balanced data, $w_c$ equals 1 for all classes as the number of instances per class is the same. Representing the true label of the data sample index $q$ as $y_q$, the weighted binary cross-entropy loss function $\mathcal{L}$ is defined as:
\begin{equation}
    \mathcal{L}(y_q,\hat{y_q}) = w_1 y_q \log(\hat{y}_q) + w_0 (1 - y_q) \log(1 - \hat{y}_q)
\end{equation}
We deploy and evaluate several image-based models using different CNN backbones as feature extractors. We train for a maximum of 100 epochs, setting a batch size of 16 per GPU and early stopping. During hyperparameter searching, Stochastic Gradient Descent (SGD) with a momentum of 0.9 and Adaptive Moment Estimation (Adam) with $\beta_1$=0.9 and $\beta_2$=0.999 are utilized. Weight decay of 0.97 is applied in both optimizers every 1k steps, whereas the learning rate ranges from 0.1 to 1e-6 in powers of 10. We also use a moving average with a 0.9999 decay rate. Two Teslas P100 GPUs with 16 GB RAM are used. 

Data augmentation is performed during training. We apply random left-right flipping at video frame level. Additionally, brightness and contrast augmentation are applied to modify the video frames' overall luminance and intensity differences. Rotations, cropping, or other spatial distortions are avoided as they might lead to the undesired effect of losing the most critical information in the thermal images, which is the presence of the newborn within the video frame.  

\subsubsection{Evaluation \& Metrics}
\label{subsubsec:metrics}

We assess precision and recall as evaluation metrics for binary experiments \cite{lipton2014optimal}. We also employ the Matthew's Correlation Coefficient (MCC) as a more comprehensive metric, defined as:
\begin{equation}
    MCC = \scalebox{1.1}{$\frac{TN\cdot TP-FP\cdot FN}{\sqrt{(TN+FN)(FP+TP)(TN+FP)(FN+TP)}}$} \vspace{6pt}
\end{equation}
where TP, TN, FP, and FN are the True Positives, True Negatives, False Positives, and False Negatives, respectively. TP occurs when the model correctly identifies positive instances, while TN denotes the correct identification of negative instances. FP arises when the model wrongly identifies negative instances as positive, and FN occurs when positive instances are wrongly identified as negative. 

Additionally, interpretability algorithms are crucial to understanding how deep learning models make decisions. To enhance system transparency and reliability, we utilize the gradient-based class activation map (GradCAM) algorithm \cite{selvaraju2017grad}. This technique generates a coarse localization map, highlighting significant regions within an image that contribute the most to predicting a specific class.

\subsection{Step III. Time of Birth Detection}
\label{subsec:tob}

In the previous step, we trained to detect frames that include a visible newborn. However, the main goal of this study is the detection of the ToB. To achieve this, we select the model with the best performance from Step II, and we perform inference on an entire video. This is done by extracting all the normalized frames $x(n)=\phi_\mathbf{v}(I(n))$ and feeding them into the model $\varphi$ with parameters $\theta$, resulting in the inferred probability scores $\hat{y}(n)=\varphi_\theta(x(n))$. 

This prediction score signal is noisy, as the individual frames are interpreted independently. To mitigate the noise, we filter the signal with a smoothing filter before predicting ToB. The chosen filter should be simple and without phase distortion, i.e. a FIR filter. As second precision is seen as precise enough for a ToB detector, we choose a filter size of $K=25$ samples, and filter coefficients $h(k)=\frac{1}{K}$, $\forall k$. The output signal (filtered score) is computed as:
\begin{equation}
    \hat{y}_{h}(n) = \sum_{k=0}^{K-1}h(k)\cdot\hat{y}(n-k)
\end{equation}
To estimate the ToB, we identify the first frame index $n_{birth}$ where the average score exceeds a predetermined confidence threshold $\gamma$. The ToB is then calculated by converting this frame index into time:
\begin{gather}
    n_{birth}=min\{n|y_h(n)\geq\gamma\} \\
    \hat{T}_{birth}=\lfloor n_{birth}\, / f_r\rfloor
\end{gather}
where $\lfloor\cdot\rfloor$ denotes the floor function and $f_r$ is the frame rate. For evaluation, we define the error $err$ as the time difference between the predicted ToB $\hat{T}_{birth}$ and the manual annotated ToB ($T_{birth}$):
\begin{equation}
    err = \hat{T}_{birth} - T_{birth}
\end{equation}
A positive $err$ indicates that the predicted ToB occurs after the actual birth, while a negative $err$ means the predicted ToB occurs before the birth. We also use the absolute error $|err|$ to compute statistical metrics such as the first quartile (Q1), median (Q2), third quartile (Q3), and mean.

\section{Experiments \& Discussion}
\label{sec:exp}

\subsection{Experimental Details}
\label{subsec:exp_details}

In this section, we introduce the experimental details for implementing binary classification of newborn visibility on thermal video frames. We first define the test set by manually selecting 10 fully annotated thermal videos, ensuring the same amount of recordings from both the delivery room and the operation theatre. This test set does not include scenarios involving twins, i.e., each thermal video contains only a single birth event. The remaining fully annotated thermal videos are allocated in the training and validation sets, performing an 85\%/15\% division.

Newborn detection is implemented as a binary classification task in order to identify the presence of the newborn within the thermal frames. However, identifying the newborn after birth is an evident difficulty, even during the annotation. The primary factor is the rapid decrease in the newborn's skin temperature right after delivery. Challenges also arise from potential obstructions to the newborn's visibility, such as being covered by towels or being handed to the mother. Given our strict criterion for annotating VNB, where clarity and visibility are prioritized, we exclude NNB instances occurring after birth. This exclusion is required as post-birth NNB instances may show a partially visible newborn. Thus, we ensure that the model learns from unambiguous instances. In addition, to mitigate the inherent imbalanced class distribution in the dataset, we implement downsampling by extracting one single frame every second of a continuous NNB sequence in a video. This approach allows us to retain NNB variability while significantly reducing the number of instances, even though full data balance is not reached. Therefore, using a weighted loss function remains necessary during the training phase. Table~\ref{tab:set_dist} summarizes the distribution of thermal video frames across train, validation, and test sets.

\begin{table}[t]
\caption{Data distribution across the different subsets. We employ downsampling on NNB to reduce the number of instances of the overrepresented class while keeping variability.}
\label{tab:set_dist}
\centering
\begin{tabular}{lccc}
\hline
\multicolumn{1}{c}{Subset} & Videos & Frames (VNB) & Frames (NNB) \\ \hline \hline
Train & 91 & 20887 & 78848 \\
Validation & 15 & 3381 & 12893 \\
Test & 10 & 1629 & 8827 \\ \hline \hline
\end{tabular}
\end{table}

After an extensive series of experiments for hyperparameter tuning, we determined that the SGD optimizer with a learning rate of 1e-4 consistently produced optimal behavior across our metrics for all the CNN backbones. We also employed transfer learning from ImageNet and fine-tuning on thermal data with no frozen layers. These choices were based on the best trade-off between convergence speed and model performance observed in the experimentation process. 

\subsection{Newborn Detection (Step II)}

\begin{table*}[t]
\caption{Performance on Step II-Newborn Detection within the thermal video using different CNN backbones. VNB is used as the positive class. The number of parameters considers both the backbone and the classifier (in millions). Depth refers to the topological depth of the network, i.e., the number of layers with parameters. This includes activation layers, batch normalization layers, etc. The average time (in milliseconds) per single inference step using only one V100 GPU is also measured.}
\label{tab:exp}
\centering
\begin{tabular}{lcccccc}
\hline
\multicolumn{1}{c}{Backbone} & Params (M) & Depth & Precision & Recall & MCC & Running Time (ms) \\ \hline \hline
MobileNetV3Small & 1.5 & 87 & 0.874 & 0.767 & 0.788 & 12.7 \\
MobileNetV3Large & 4.0 & 109 & 0.715 & 0.853 & 0.736 & 16.5 \\
EfficientNetV2B0 & 7.2 & 151 & 0.836 & 0.918 & 0.852 & 20.4 \\
\textbf{EfficientNetV2B1} & 8.2 & 185 & \textbf{0.881} & 0.893 & \textbf{0.866} & 23.6 \\
EfficientNetV2B2 & 10.2 & 193 & 0.848 & 0.919 & 0.86 & 24.5 \\
EfficientNetV2B3 & 14.5 & 225 & 0.84 & \textbf{0.924} & 0.858 & 28.7 \\ \hline
\end{tabular}
\end{table*}

We present the main results regarding newborn detection within thermal video frames in Table~\ref{tab:exp}. The performance comparison across various CNN backbones reveals interesting insights. Firstly, the increased model complexity and architecture optimization of EfficientNetV2 contributes to a more effective representation of the underlying patterns in the data while maintaining comparable computational efficiency to MobileNetV3. Secondly, the different ``Base'' versions of EfficientNetV2 provide similar metrics, so the selection entails a trade-off between model size, performance, and efficiency. In our study, EfficientNetV2B1 stands out as the top-performing model, surpassing other backbones in Precision and MCC without requiring high computational costs.

\begin{figure*}[!ht]
\centering
\begin{minipage}[b]{.8\linewidth}
  \centering
  \centerline{\includegraphics[width=\linewidth]{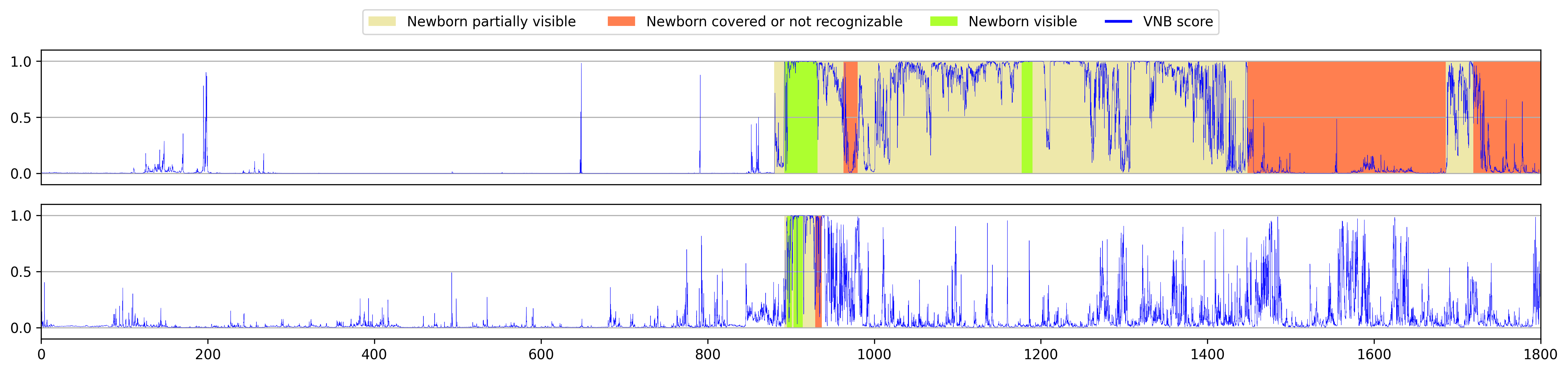}}
%  \vspace{1.5cm}
\end{minipage}
\caption{Illustration of prediction scores from thermal videos in Step II-Newborn Detection. The prediction score of VNB is inferred at every thermal frame by our EfficientNetV2B1 model with GMM normalization. The x-axis describes the duration of the thermal videos in seconds (around 30 minutes). The y-axis represents the probability of VNB between 0 and 1. The white background indicates the absence of the newborn within the video. In the delivery room (top row), the newborn is typically handed to the mother immediately after birth and covered with towels, resulting in partial visibility of the newborn in most post-birth frames. In the operation theatre (bottom row), the newborn is commonly visible for a short period of time as midwives relocate the newborn to a warmer environment beyond the camera's field of view after umbilical cord clamping. Moreover, the abdomen of the mother is often visible during surgery, potentially leading to an increased number of FP before and after birth.}
\label{fig:timeline}
\end{figure*}

Our newborn detector shows promising results in identifying TP around the manually annotated ToB, as depicted in Figure~\ref{fig:timeline}. The primary reason for the majority of FP occurring before birth is the existence of warm elements in the thermal frame. This situation is particularly observed in the operation theatre, the bottom timeline of the figure, where the visibility of the mother's abdomen during surgery reveals elevated temperatures, leading the model to make inaccurate predictions. In the delivery rooms, scenarios where the newborn faces troubles passing through the birth canal may also lead to FP right before the birth. On the other hand, FP are expected to occur after birth. As mentioned, this is mainly caused by the implementation of a rigorous VNB annotation criterion and, as a result, the need to exclude noisy NNB instances after birth during training to avoid model misinterpretations.

\begin{figure*}[!ht]
\centering
\begin{minipage}[b]{.85\linewidth}
  \centering
  \centerline{\includegraphics[width=\linewidth]{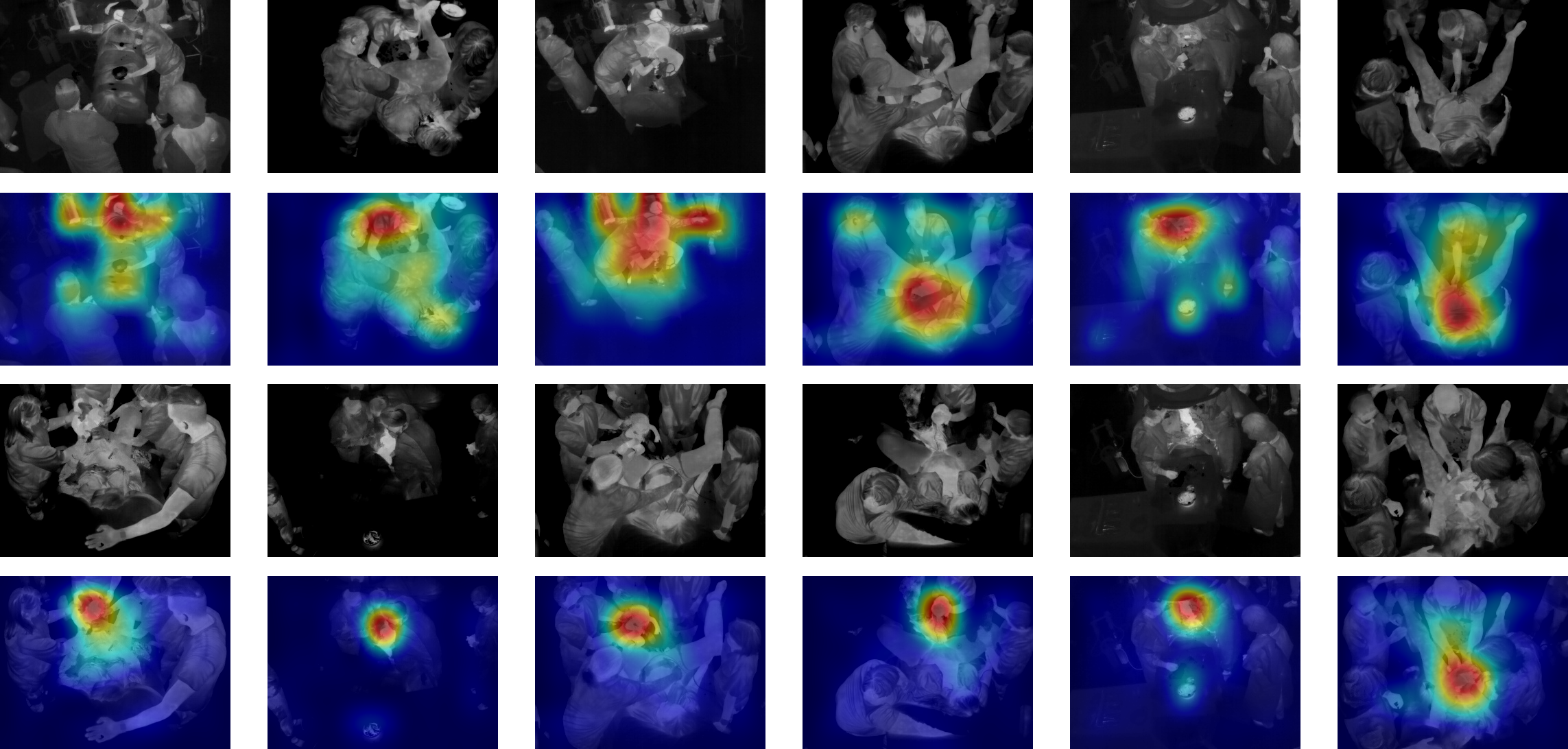}}
%  \vspace{1.5cm}
\end{minipage}
\caption{Visualization of the GradCAM algorithm on the output of Step II-Newborn Detection. Thermal video frames are normalized with our proposed GMM normalization and passed through our trained EfficientNetV2B1 model. First and second rows: examples of NNB. Third and fourth rows: examples of VNB.}
\label{fig:gradcam}
\end{figure*}

In order to make the decision-making processes more transparent for newborn detection, we incorporate GradCAM visualizations extracted from the last convolutional layer of the trained EfficientNetV2B1. In Figure~\ref{fig:gradcam}, this interpretability algorithm shows that our system can identify the most significant areas in the thermal video frames. A noteworthy observation is the model's adeptness in broadening its focus when the newborn is not visible within the video frame, contrasting with its ability to concentrate on a more specific region when the newborn is present. The impact of the GMM normalization is also evident in these visualizations, where background information is largely eliminated, and the model is guided to focus on the most relevant pixels for our task.

Additionally, Table~\ref{tab:abl_study} presents results from supplementary experiments conducted using the EfficientNetV2B1 backbone. Here, we analyze the effectiveness of our proposed GMM normalization compared to the conventional Max-Min technique. Analyzing the top models from this table, GMM outperforms Max-min in terms of precision and MCC. This is advantageous for estimating the ToB, as our goal is to minimize FP. Moreover, we evaluate the implications of using pre-trained weights when transitioning from RGB images (as employed in ImageNet) to single-channel grayscale images typical of thermal imaging. The outcomes demonstrate that transfer learning accelerates model convergence, enhances generalization, and effectively addresses challenges associated with limited data. We also explore the influence of modifying the luminance and intensity values of the thermal frames during the data augmentation process. It is noteworthy that augmentation barely improves results when using Max-Min normalization. This technique, which normalizes each thermal frame based on its maximum and minimum values, may introduce varying luminance levels across frames within the same video, resembling a form of data augmentation. Additional brightness and augmentation may aggravate pixel value saturation in high-intensity values, causing the loss of the most fundamental information. In contrast, our GMM normalization employs a predefined range that ensures the capture of human skin temperature while allowing sufficient flexibility to cover higher temperatures. Consequently, luminance and intensity data augmentation prove more advantageous for GMM than Max-Min normalization, mitigating the risk of saturation of high-intensity values and facilitating diverse variations with fewer saturation-induced distortions.

\begin{table}[t]
\caption{Supplementary experiments using EfficientNetV2B1 backbone in Step II-Newborn Detection. Norm.: Normalization technique. Aug.: Data augmentation. Weights: ImageNet pre-trained weights.}
\label{tab:abl_study}
\centering
\begin{tabular}{lccccc}
\hline
\multicolumn{1}{c}{Norm.} & Aug. & Weights & Precision & Recall & MCC \\ \hline \hline
GMM & - & - & 0.499 & 0.565 & 0.438 \\
GMM & \ding{51} & - & 0.073 & 0.006 & 0.025 \\
GMM & - & \ding{51} & 0.798 & 0.892 & 0.813 \\
\textbf{GMM} & \ding{51} & \ding{51} & \textbf{0.881} & 0.893 & \textbf{0.866} \\
Max-Min & - & - & 0.419 & 0.822 & 0.482 \\
Max-Min & \ding{51} & - & 0.252 & 0.252 & 0.113 \\
Max-Min & - & \ding{51} & 0.817 & \textbf{0.92} & 0.841 \\
Max-Min & \ding{51} & \ding{51} & 0.838 & 0.918 & 0.853 \\ \hline \hline
\end{tabular}
\end{table}

\subsection{Time of Birth Detection (Step III)}

\begin{figure}[t]
\centering
\begin{minipage}[b]{\linewidth}
  \centering
  \centerline{\includegraphics[width=\linewidth]{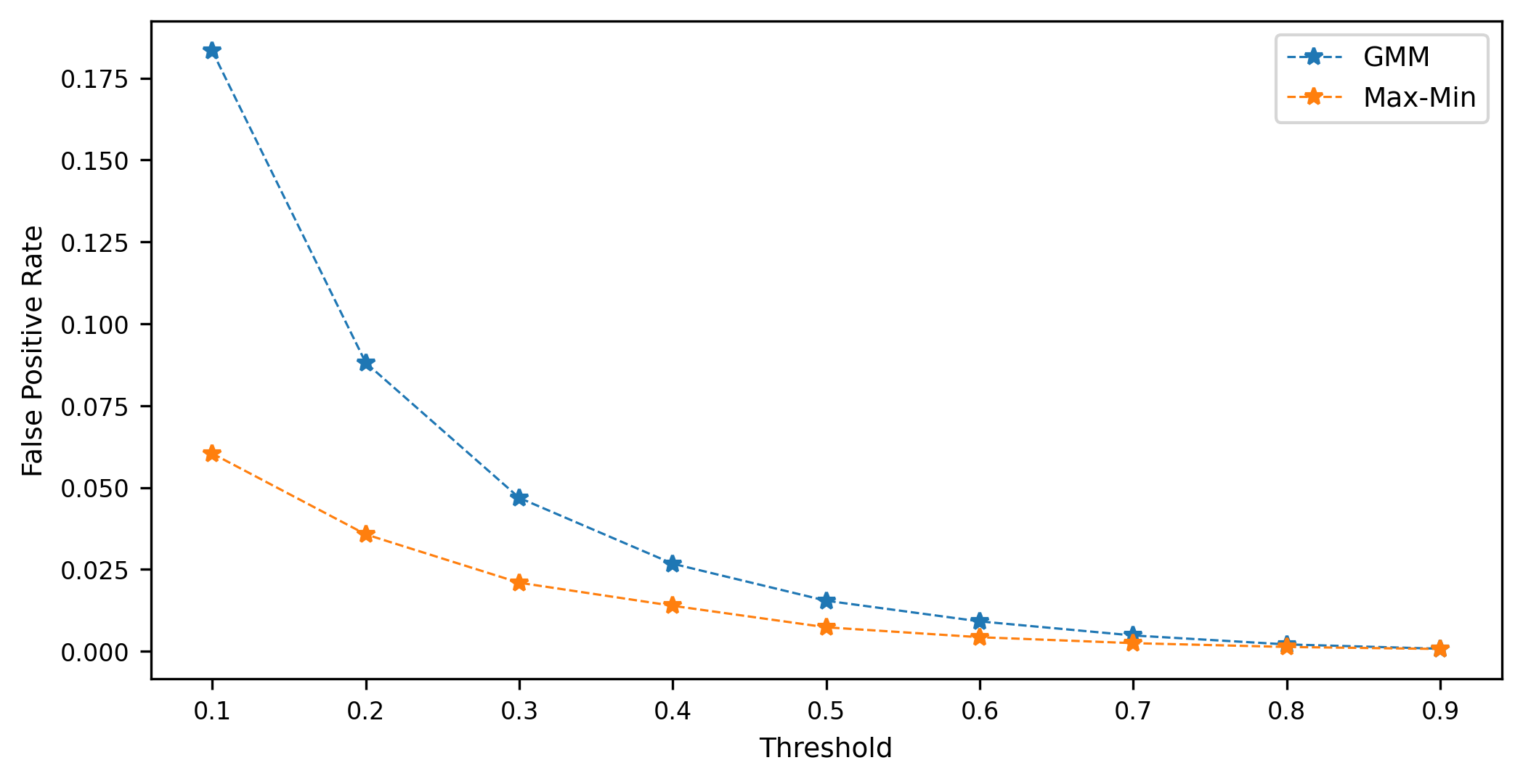}}
%  \vspace{1.5cm}
\end{minipage}
\caption{Evaluation of FPR at various confidence thresholds. For high-confidence thresholds, both techniques exhibit superior capabilities in identifying the newborn, leading to more reliable and accurate outcomes in estimating the ToB.}
\label{fig:fpr-mcc}
\end{figure}

Given that our system relies on accurately identifying the presence of the newborn within a fixed timeframe for estimating the ToB, minimizing FP instances before birth is critical. Therefore, the confidence threshold $\gamma$ for estimating the ToB is selected based on the capability of a model to mitigate FP while maintaining a satisfactory performance. To achieve this, we first filter the raw prediction scores $\hat{y}$, followed by an examination of the filtered scores for all the thermal frames manually annotated as NNB (before birth) and VNB from the test set. In particular, we evaluate the models that exhibit the highest MCC score in Table~\ref{tab:abl_study} when comparing GMM and Max-Min normalization techniques. Figure~\ref{fig:fpr-mcc} illustrates the False Positive Rate (FPR) of $\hat{y}_h(t)$ at various thresholds. The best performance is observed when $\gamma=0.9$ for both normalization techniques. 

\begin{figure*}[t]
\centering
\begin{minipage}[b]{\linewidth}
  \centering
  \centerline{\includegraphics[width=.99\linewidth]{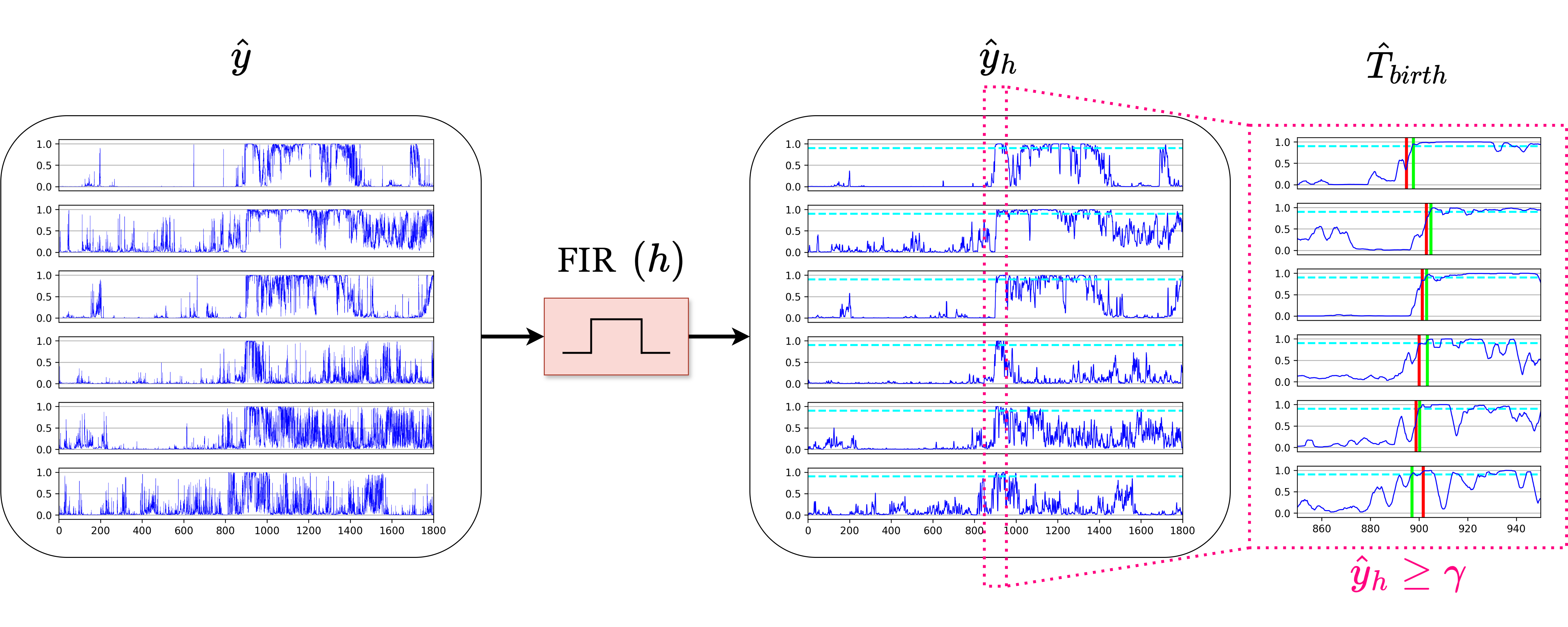}}
%  \vspace{1.5cm}
\end{minipage}
\caption{Visualization of the post-processing stage for estimating the ToB. A sliding window $h$ is used to average (filter) the raw prediction scores $\hat{y}$ computed in Step II-Newborn Detection. These filtered scores $\hat{y}_h$ are subsequently used in Step III-ToB Detection to find the first frame index that exceeds a predefined confidence threshold $\gamma=0.9$, marked in cyan, and then estimate the ToB. In the rightmost graph, the estimated ToB is depicted in green. The manually annotated ToB is highlighted in red.}
\label{fig:filter}
\end{figure*}

\begin{figure}[!ht]
\centering
\begin{minipage}[b]{.8\linewidth}
  \centering
  \centerline{\includegraphics[width=\linewidth]{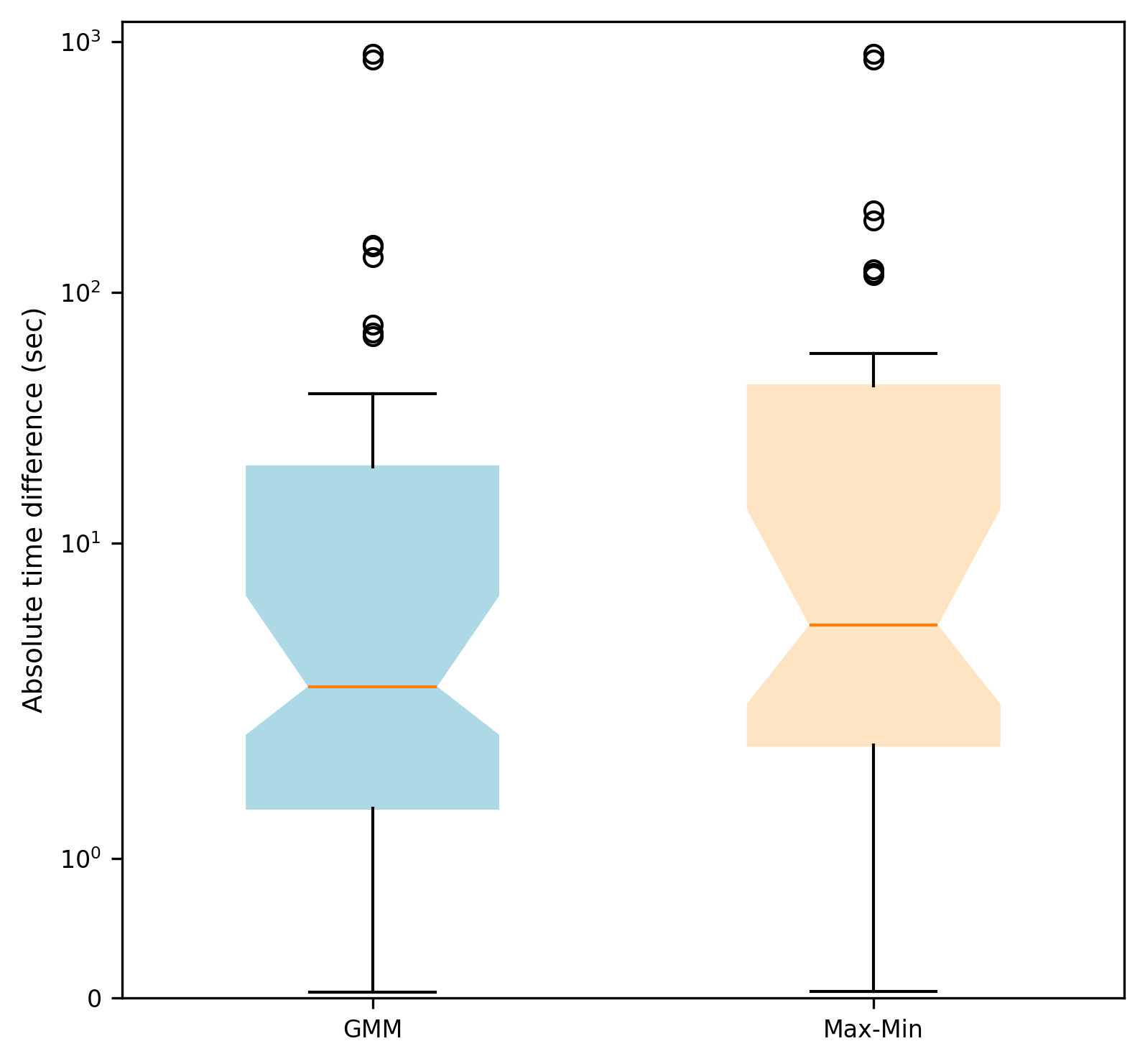}}
%  \vspace{1.5cm}
\end{minipage}
\caption{Boxplot representation of the distribution of the absolute time difference between the manually annotated ToB and the estimated ToB. The notch-like indentation represents a confidence interval of 95\% around the median. Whiskers are set at the lowest and highest datum within the interval [Q1-1.5\*IQR, Q3+1.5\*IQR], with IQR=(Q3-Q1) denoting the interquartile range. The y-axis represents the time difference in seconds using an exponential scale. The x-axis indicates the two normalization methods to be compared.}
\label{fig:boxplot}
\end{figure}

\begin{table}[!ht]
\caption{Statistical performance in Step III-ToB Detection. A comparison between GMM and Max-Min normalization techniques in estimating the ToB is presented. All values are provided in seconds. Q values represent the quartiles. In both models, $\hat{T}_{birth}$ was found in 100\% of the test videos.}
\label{tab:boxplot}
\centering
\begin{tabular}{lccccc}
\hline
\multicolumn{1}{c}{Normalization} & Q1 & Q2 & Q3 & Mean \\ \hline \hline
GMM & 1.4 & 2.7 & 20.2 & 69.8 \\
Max-Min & 1.8 & 4.7 & 42.4 & 76.8 \\ \hline \hline
\end{tabular}
\end{table}

This confidence threshold, along with the filtered scores $\hat{y}_h$, are used to estimate the ToB using our proposed approach outlined in Section~\ref{subsec:tob}. Examples of this process are illustrated in Figure~\ref{fig:filter}. To ensure a more equitable comparison between GMM and Max-Min normalization techniques, we estimate the ToB from a more representative subset of data. This new subset comprises 36 thermal videos, including the 10 videos from the test set and the 26 additional non-fully annotated videos containing solely ToB information. A graphical distribution of the absolute time difference between the estimated ToB and the manually annotated ToB is illustrated in Figure~\ref{fig:boxplot}. These results demonstrate a more favorable outcome when employing GMM normalization. From Table~\ref{tab:boxplot}, we observe that GMM achieves a median value closer to the optimal absolute difference (0 seconds), providing more accurate ToB estimations than Max-Min normalization.

These results hold great promise for developing a real-time ToB detector. Notwithstanding this, whereas this system may provide reasonably accurate results in many thermal videos, it becomes challenging in certain scenarios. As mentioned earlier, the presence of warm elements can lead to inaccurate predictions of VNB. Additionally, situations where the newborn is experiencing difficulty passing through the birth canal, only partially visible, or hidden due to factors such as birthing position, healthcare provider's location, or occlusions may contribute to inaccuracies in detecting the newborn within the thermal frame and, therefore, assessing the ToB. Further research is necessary to address these complexities in the ToB estimation.

\section{Conclusion}
\label{sec:R&D}

In this work, we introduce the first AI-driven ToB detector based on thermal imaging. The inherent challenges of working with thermal data make reliance on absolute temperature values impractical. Moreover, difficulties arise when attempting to normalize thermal frames both within a single video and across multiple videos in a multi-camera setup. To address these adversities, we propose an adaptive normalization approach based on GMM, aiming to generate more consistent relative temperature measurements around the human skin temperature. Using the EfficientNetV2B1 backbone fine-tuned on thermal birth video frames as a feature extractor, our system achieves a precision of 88.1\% and a recall of 89.3\% in detecting the presence of the newborn within thermal video. Furthermore, by postprocessing the prediction scores, our system accurately estimates ToB with an absolute median deviation of 2.7 seconds relative to the manual annotations. 

While the obtained results offer promising implications for automated ToB detection, this study has some limitations. Our system exclusively relies on the spatial information of the thermal frames to estimate the ToB, omitting valuable insights from the temporal dimension. Besides, our proposed GMM normalization is done after the full thermal video is generated. While this approach is suitable for debriefing, quality improvement, and research purposes, it may not be adequate for real-time decision support. In such scenarios, it becomes crucial to locate the ToB in real time and periodically update the parameters of the GMM based on the available thermal frames.

Future search efforts will prioritize addressing these limitations to enhance ToB estimates, focusing on the analysis of thermal video clips. Furthermore, extending our research to refine our proposed GMM normalization approach or exploring alternative normalization techniques will deepen our understanding of thermal data standardization. Lastly, we will integrate the ToB detector with our previously developed NRAA classifier to culminate the creation of NewbornTimeline --- a complete system capable of automatically generating detailed NRAA timelines, including activities and events during and after birth.

\section*{Funding}

The NewbornTime project is funded by the Norwegian Research Council (NRC), project number 320968. Additional funding has been provided by Helse Vest, Fondation Idella, and Helse Campus, Universitetet i Stavanger. Study registered in ISRCTN Registry, number ISRCTN12236970.

\section*{Acknowledgements}

The study has been approved by the Regional Ethical Committee, Region West, Norway (REK-Vest), REK number: 222455. The project has been recommended by Sikt - Norwegian Agency for Shared Services in Education and Research, formerly known as NSD, number 816989. Informed consent was obtained from all mothers involved in the study.

We would like to express our gratitude to all mothers, healthcare providers, and mercantile personnel who made this study possible. Our gratitude also goes to all contributors involved in the NewbornTime project.

\vfill\pagebreak

\bibliographystyle{cas-model2-names}
\bibliography{main}

\end{document}